\documentclass[journal]{journal}
\usepackage{graphicx}
\usepackage{microtype}
\usepackage{graphicx}
\usepackage{subfigure}
\usepackage{booktabs} 
\usepackage{wrapfig}
\usepackage{float}
\usepackage{array}
\usepackage{amssymb}
\usepackage{pgfplots}
\pgfplotsset{compat=1.8}
\usepgfplotslibrary{statistics}
\graphicspath{  {./images/}  }
\usepackage{multirow}

\ifCLASSINFOpdf
\else
\fi
\begin{document}
%
\title{Self-Supervised Pretraining on Paired Sequences of fMRI Data for Transfer Learning to Brain Decoding Tasks}
%
%
%

\author{Sean~Paulsen,
        and~Michael~Casey
        
\thanks{S. Paulsen is with the Department
of Computer Science, Dartmouth College, Hanover,
NH, 03755 USA e-mail: paulsen.sean@gmail.com}
\thanks{M. Casey is with the Department of Music and Department of Computer Science,  Dartmouth College}%
\thanks{Manuscript received May 4, 2023.}}

%
%

\markboth{International Conference on Pattern Recognition, Machine Learning and Consciousness}%
{Paulsen and Casey}
%



\maketitle
\thispagestyle{empty}

\begin{abstract}
In this work we introduce a self-supervised pretraining framework for transformers on functional Magnetic Resonance Imaging (fMRI) data. First, we pretrain our architecture on two self-supervised tasks simultaneously to teach the model a general understanding of the temporal and spatial dynamics of human auditory cortex during music listening. Our pretraining results are the first to suggest a synergistic effect of multitask training on fMRI data. Second, we finetune the pretrained models and train additional fresh models on a supervised fMRI classification task. We observe significantly improved accuracy on held-out runs with the finetuned models, which demonstrates the ability of our pretraining tasks to facilitate transfer learning. This work contributes to the growing body of literature on transformer architectures for pretraining and transfer learning with fMRI data, and serves as a proof of concept for our pretraining tasks and multitask pretraining on fMRI data. 
\end{abstract}


%
\IEEEpeerreviewmaketitle

\section{Introduction}
%
%
%
%
\IEEEPARstart{F}{unctional} MRI (fMRI) scans measure blood-oxygen-level-dependent (BOLD) responses that reflect changes in metabolic demand consequent to neural activity \cite{bedel, hillman2014, rajapakse1998}. By measuring BOLD responses at specific combinations of spatio-temporal resolutions and coverages, fMRI data provide the means to study complex cognitive processes in the human brain \cite{kubicki2003, wang2005, papma2017}. In particular, task-based fMRI protocols include targeted stimuli or other task variables, such as question answering, during the scan. Researchers can then conclude associations between task features and the evoked responses across the brain \cite{li2009, venkataraman2009, nishimoto2011}. Regions of activity that are correlated with the presence of a particular task feature are thus taken to be involved in the brain's representation of that feature \cite{simon2004}, and they are considered to be functionally connected \cite{rogers2007}. Even rest-state fMRI data, that is, data collected in the absence of external stimuli or task, contain characteristic multi-variate signals of the brain \cite{audimg, niu2021, yeo2011, vandijk2010, hu2006interregional}. Such rest-state signals have been shown to be predictive of the diagnosis and characterization of multiple neurological diseases and psychiatric conditions \cite{zhan2015, woodward2015, xia2018}.

fMRI researchers have thus adopted machine learning (ML) techniques to analyze the complex relationship between BOLD signal and the underlying task, disease, or biological information. More specifically, training an ML model to predict such information given the BOLD data as input is known as \textbf{task-state decoding}, or \textbf{brain decoding}. Toward the goal of more powerful brain decoding models, many advances in modern \textit{deep} machine learning have been applied to fMRI research. These include convolution-based models \cite{zou2017, kawahara2017, audimg}, recurrent neural networks (RNN) \cite{dakka2017}, and graph neural networks \cite{li2021gnn}. Most recently, transformer \cite{vaswani2017} based models have achieved state of the art results on several brain decoding tasks \cite{malkiel, bedel, nguyen}, having already grown to dominate the fields of time series forecasting \cite{li2020timeseries}, natural language processing \cite{devlin2019}, and computer vision \cite{dosovitskiy2021, li2019visual}.

However, training deep models is data intensive, while fMRI scans are expensive with relatively little data obtained per scan. One strategy to somewhat alleviate the burden of data is to \textbf{pretrain} the model on a \textbf{self-supervised} task to acquire general knowledge inherent in the dataset. The pretrained model then has a head start, so to speak, on the task of interest, by leveraging its general understanding of the data \cite{Erhan}. This strategy is nearly ubiquitous in the domain of Natural Language Processing (NLP) \cite{kalyan2021} and has begun to appear in fMRI studies \cite{nguyen,malkiel}. As Kalyan et al. (2021) \cite{kalyan2021} note, ``These models provide good background knowledge to downstream tasks which avoids training of downstream models from scratch." This process is called \textbf{transfer learning}. In this paper we propose two new self-supervised pretraining tasks on sequences of audio-evoked fMRI data to facilitate transfer learning to downstream auditory brain decoding tasks. We demonstrate our transformer architecture's ability to learn these tasks and ``transfer" that knowledge to improve convergence time on a supervised auditory brain decoding task. Further, our results show that simultaneous training on both pretraining tasks achieves superior final performance than training on only one of the tasks.

Our contributions are: (\textbf{1}) we present novel self-supervised tasks for two-task simultaneous pretraining on sequences of fMRI data, (\textbf{2}) we report our transformer architecture's successful learning of those tasks and achieve, to the best of our knowledge, the first evidence of a synergistic benefit from multitask training on fMRI data, (\textbf{3}) we demonstrate transfer learning to a supervised brain decoding task, and thereby establish a proof of concept of our pretraining tasks' suitability and our framework's capacity for transfer learning on fMRI data.

\section{Related Work}
Univariate approaches to fMRI data such as contrast subtraction can be useful for basic analysis, but such approaches struggle to isolate the densely overlapping patterns of multivariate signals which comprise neural activity \cite{penny2011, woolrich2001}. This challenge motivated the adoption of early ML architectures for multivariate fMRI analysis \cite{norman2006, haxby2012}, notably support vector machines for brain decoding classification \cite{imagined, song2014, wang2007, hojjati2017}. Progression into \textit{deep} ML models saw multilayer perceptrons \cite{suk2016}, autoencoders \cite{audimg, huang2018}, convolutional neural networks (CNN) \cite{wang2020, yamins2014}, and graph neural networks (GNN) \cite{li2021gnn} for feature extraction and classification of single fMRI images. Time series analysis is perhaps more desirable due to the high degree of temporal correlation in BOLD responses, and indeed recurrent neural networks (RNN) and various long short-term memory (LSTM) models have been reported \cite{dvornek2017, li2020lstm, zhao2020, thomas2019, pominova2018}.

Most recently, the transformer \cite{vaswani2017} architecture has emerged as a superior alternative to recurrent methods for fMRI timeseries modeling. Bedel et al. (2023) \cite{bedel} improved the state of the art for timeseries classification on multiple public fMRI datasets with a novel fused-window attention mechanism, but their work did not explore pretraining or transfer learning.  Nguyen et al. (2021) \cite{nguyen} achieved state of the art classification accuracy for a task-state decoding task on the Human Connectome Project 7-task dataset\cite{hcp}. Their analysis includes the explicit benefits of the transformer's self-attention module when compared to previous recurrent architectures, as well as a demonstration of transfer learning when pretraining on held-out subsets of HCP 7-task. However, their pretraining task was supervised classification specific to HCP 7-task labelled data, and thus their pretrained models would be of little to no value toward transfer learning on different datasets or modalities \cite{kalyan2021}. Malkiel et al. (2022) \cite{malkiel} pretrain on a self-supervised fMRI reconstruction task by wrapping the transformer block in an encoder-decoder. They report that their pretraining was crucial for improved state of the art performance on a variety of fMRI tasks such as age and gender prediction, and schizophrenia recognition. We note that their downstream task uses the CLS token decoding method popularized by Devlin et al. (2019) \cite{devlin2019}, while their pretraining task does \textit{not} incorporate the CLS token. This inconsistency between training phases does not obtain the full value of the transfer learning paradigm.

Extending the above work, we explore multitask pretraining and transfer learning with novel self-supervised pretraining tasks which include the CLS token, with all model inputs in a standardized geographical brain space, \textit{without} passing through an embedding layer.

\begin{figure}[t]

    \centering
    \includegraphics[width=0.3\textwidth]{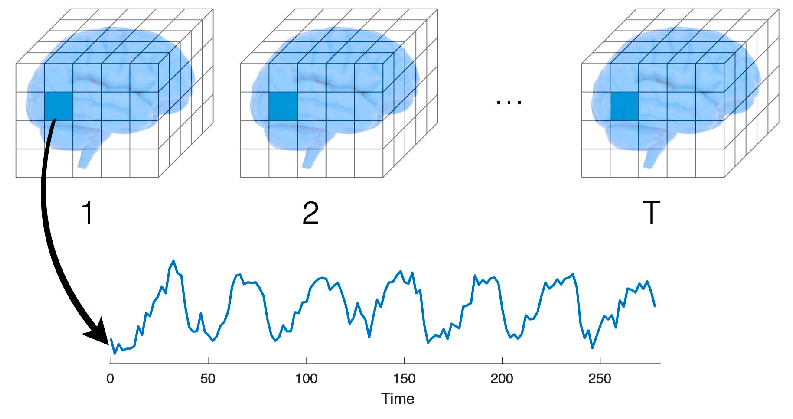}
    \caption{The sequences of voxel data used in our experiments are timeseries of neural activity measured by fMRI. Graphic published in \cite{voxelfig}}
    \label{fig:voxels}
\end{figure}

\begin{figure*}[t]

    \centering
    \includegraphics[width=1.0\textwidth]{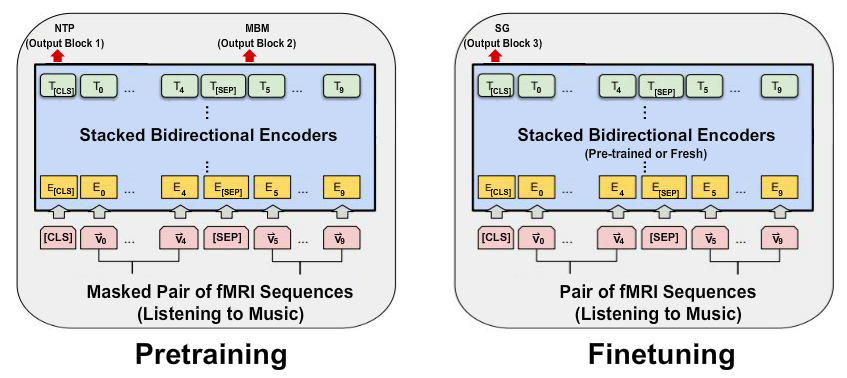}
    \caption{Pretraining and Finetuning phases. Output Blocks are not pictured but are detailed in corresponding sections. The model learns to extract information into the CLS token, which is fed to Output Block 1 during pretraining, and Output Block 3 during finetuning, for classification. The SEP token separates the two sequences. The masked token(s) are fed to Output Block 2. For finetuning, the model either loads the pretrained weights or trains a fresh model. In either case, all parameters are trained.}
    \label{fig:architecture}
\end{figure*}

\section{Architecture and Training Tasks} 
\subsection{Paired-Sequence Transformer}
Our architecture is a modified stacked bidirectional-encoder design (Figure \ref{fig:architecture}) with two separate output blocks, one for each of two self-supervised pretraining tasks on which the model is trained simultaneously. We implemented our models from scratch with the pyTorch library. Our model does not include the standard embedding layer after positional encoding. We hypothesize that the composition of the fMRI scanner's measurement of BOLD signal with the mapping of that measurement to MNI space constitutes a meaningful embedding of the physical, biological neural response. The data are \textit{already} in a shared, distributed, representative space. Hence, we dispense with the embedding layer in our design.

A thorough explanation of the data preprocessing and the construction of the inputs to the model is given in the Data Preparation section below, which we summarize here. All training data in this work were built from the Music Genre fMRI Dataset (2021) \cite{opengenre}. The images were collected while five subjects listened to samples of music from ten different genres. Each input to the model is constructed by extracting a contiguous sequence of five fMRI images of a subject listening to music (Seq1), and pairing it with another (different) such sequence (Seq2). A \textbf{separator token} (SEP) \cite{devlin2019} is inserted between the two sequences, and a \textbf{classification token} (CLS) \cite{devlin2019, malkiel, nguyen} is inserted at the front. Before constructing the inputs, we reduced all fMRI images to only the left-side auditory cortex, resulting in 420 voxels, which we then flattened to 1-D. Thus each input $x_i$ in the training set is a sequence of twelve 420-dimensional vectors:
\begin{equation}
    x_i = \left[ CLS, \: \vec{v}_{0}, \: \ldots ,\: SEP, \: \vec{v}_{5} \: \ldots,\: \vec{v}_{11} \right],\;\;\; v_j \in \mathbb{R}^{420}.
\end{equation}

The implementation of these tokens without an embedding layer is explained in the last paragraph of the Materials section below.

\subsection{Pretraining Tasks} 
We now present our two novel self-supervised pretraining tasks. First is \textbf{Next Thought Prediction} (NTP). The goal of this task is binary classification, predicting whether or not Seq2 follows immediately after Seq1 in the original data. From the output of the final transformer block, the transformed CLS token is sent to \textbf{Output Block 1}. This block consists of a linear layer projecting down from 420 dimensions to 210, then a linear layer projecting down from 210 to 2, and finally a softmax is applied to obtain probabilities for ``No" (index 0) and ``Yes" (index 1). The loss for NTP is calculated as the Cross-Entropy between the result of Output Block 1 and a one-hot encoding of the ground truth.

Our second pretraining task is \textbf{Masked Brain Modeling} (MBM). The goal of this task is to reconstruct a masked element or elements of the input sequence. When an input arrives at the model, before positional encoding, either one or two of the ten fMRI images are chosen uniformly at random (without replacement) for masking. It is a 50/50 chance whether one or two are chosen. When an image is chosen, there is an 80\% chance to replace it with the \textbf{mask token} (MSK), a 10\% chance to replace it with a random image sampled uniformly from the full dataset, and a 10\% chance to leave it unchanged. The chosen indices are recorded, and the elements of the final transformer block's output at those indices are passed separately to \textbf{Output Block 2}. This block consists of a dense layer with ReLu activation, then a second dense layer with linear activation. The loss for MBM is calculated as the Mean Squared Error between the output and the original chosen fMRI image. In the case of two chosen images, the total MBM loss is the average of the two individual MBM losses.    

Note the inherent data augmentation of the MBM task. There are ten fMRI images in each training sample, and the result of each possible masking configuration yields a distinct training sample. Thus MBM can effectively grow the size of the training set by an order of magnitude if the model is trained long enough. This gain is perhaps overlooked in domains such as natural language processing where billions of training samples are available. In fMRI studies, however, data poverty is a consistent concern due to the financial and time costs of scanning. While we make no specific claims about the effects of this augmentation in this work, this potential benefit built into the task is noteworthy. 

\subsection{Multitask Learning}
Training on more than one task simultaneously, known as \textbf{Multitask Learning} (MTL), has been shown to improve downstream performance in several domains \cite{mtl} by benefiting from the underlying \textit{relationships} between tasks, but to our knowledge this has not previously been done when training on fMRI data. In their thorough treatment of the brain's musical reward system, Salimpoor et al. (2015) \cite{musicpleasure} comment ``music pleasure is thought to rely on generation of expectations, anticipation of their development and outcome, and violation or confirmation of predictions.'' In other words, the notion of ``what comes next'' is intimately connected to the explicit values of voxel activity. NTP and MBM embody these two concepts, so indeed our multitask pretraining scheme is aligned with the literature. 

The raw loss value of NTP for a single training sample is, on average, at least an order of magnitude greater than the loss value of MBM when training begins. Therefore the parameter updates will certainly be dominated by NTP, stifling any learning from MBM. Deriving a theoretically optimal way to combine the two losses would be a significant endeavor, so our total training loss for a single sample is merely the weighted sum of our two loss values: \begin{equation}
    E_{multi} = \alpha_1*E_{NTP} + \alpha_2*E_{MBM}
\end{equation}
\begin{equation}
    \alpha_1+\alpha_2 = 1.
\end{equation}
These two weights are simply hyperparameters to be tuned. We explore these and other hyperparameters in the Experiments and Results section below.

\subsection{Finetuning Task}
Our novel supervised brain decoding task used for finetuning is the \textbf{Same Genre} (SG) task. The goal of this task is binary classification, predicting whether or not Seq1 corresponds to listening to the same genre of music as Seq2. From the output of the final transformer block, the transformed CLS token is sent to Output Block 3. This block consists of a single linear layer projecting down from 420 dimensions to 2, then a softmax is applied to obtain probabilities for ``No" (index 0) and ``Yes" (index 1). The loss for SG is calculated as the Cross-Entropy between the result of Output Block 3 and a one-hot encoding of the ground truth. We made the finetuning output block as simple as possible to ensure that the brunt of the work is supported by the pretrained transformer blocks.

\section{Experiments and Results}
\subsection{Hyperparameter Search}
One of the most important questions to ask in the context of multitask learning is whether the model would have been better off with only one task. In particular, \textit{how much} is the performance on NTP impeded by having to learn MBM at the same time? To explore this, we performed our hyperparameter grid search for training on the multitask regimen as well as NTP alone. Table~\ref{pthpsearch} shows the best performing (i.e. achieved the highest validation accuracy on NTP at some point during training) configurations. We let the NTP task guide our search because its binary accuracy is simply more interpretable than any metric involving the MBM task. Nevertheless, the multitask models' performance on MBM is included in our analysis below.  

All training during grid search held out run 0 from the dataset as a validation split. We applied a dropout rate of 0.1 in all transformer blocks. Models were trained for ten epochs via backpropagation with the Adam optimizer with $\beta_1=0.9, \beta_2=0.999,$ and weight\_decay $=0.0001$. 

\begin{table}[t]
\caption{Best performing configuration for the two training regimens. Parameters from top to bottom are the alpha weights for loss calculation, learning rate, number of attention heads, and factor of forward expansion in the encoder blocks.}
\label{pthpsearch}
\vskip 0.15in
\begin{center}
\begin{small}
\begin{sc}
\begin{tabular}{lcccr}
\toprule
 & Multi & NTP \\
\midrule
$\alpha_1,\alpha_2$    & 0.1, 0.9 & N/A \\
LR & $10^{-4}$ & $10^{-5}$ \\
Atn Hds    & 2& 2 \\
F Exp    & 4& 4  \\

\bottomrule
\end{tabular}
\end{sc}
\end{small}
\end{center}
\vskip -0.3in
\end{table}

In general, fewer attention heads with more layers outperformed the reverse. It is reassuring to obtain the same value for attention heads and forward expansion on both regimens. The best performing learning rate for NTP-only is an order of magnitude smaller than for multitask, but this is not surprising. NTP's contribution to the loss is scaled by $\alpha_1=0.1$, and in the most basic gradient descent, scaling the loss function by a constant is functionally the same as scaling the learning rate by that constant instead. The Adam optimizer is a bit more complex, but the general idea holds. 

\subsection{Pretraining} 
After identifying the best performing hyperparameters for both cases, we performed 12-fold cross validation for both multitask and NTP-only, where each fold holds out one of twelve runs from the dataset. It was unclear during hyperparameter search whether a 3 or 4 layer model was superior, so we considered both here. The same Adam specifications as hyperparameter search were used. The exact details of pretraining dataset construction can be found in the Materials section below, but we note here that each fold has 10,000 training samples and 800 validation samples.

Results are presented in Table~\ref{table:pretraincrossval}. For each fold, we saved the model's state after the epoch with the highest NTP accuracy on the validation split (``Best Val Acc" in the table). The ``Best Epoch" column contains the epoch in which this accuracy was achieved. The MBM loss calculated on the validation split after the Best Epoch is also given to consider the relationship between the two tasks. Consider as a baseline that the MBM training loss on the first training sample seen by a model is around $0.4$. The averages of each column are given in the last row of the table.

\begin{table*}[t]
    \begin{center}
    \caption{Results of 12-fold cross validation for Multitask (NTP and MBM) and NTP-only pretraining regimens, on 3 and 4 layers. Best Val Acc is the highest accuracy obtained during training on the NTP task on the validation split. The epoch in which that accuracy was obtained is given in the Best Epoch column, from 0 to 9 inclusive. MBM Loss is the loss obtained on the MBM task on the validation split in the Best Epoch. The average across all twelve folds is given at the bottom of each column.}
    \label{table:pretraincrossval}
    \begin{tabular}{|| m{2.2cm}| m{1.5cm} | m{2.8cm} | m{1.5cm} | m{2cm} | m{2.8cm} | m{1.5cm} ||}
    \hline
    & & \multicolumn{3}{|c|}{\textbf{Multitask (NTP and MBM)}} & \multicolumn{2}{|c||}{\textbf{NTP Only}}\\
    \hline
    \textbf{Heldout Run} & \textbf{N Layers} & \textbf{Best Val Acc} & \textbf{Best Epoch}   & \textbf{MBM Val Loss}  & \textbf{Best Val. Acc} & \textbf{Best Epoch}\\
    \hline
    \multirow{ 2}{*}{0} & 4 & 93.5\% & 8 & 0.00103  &  88.375\%   &  6   \\
    & & & & & & \\[-1em]
    & 3 & 88.125\% & 8 & 0.00048 & 88.25\% & 9\\
    \hline
    \multirow{2}{*}{1} & 4 & 87.375\% & 9 & 0.00088  &  87.375\%   &  8   \\
    & & & & & & \\[-1em]
    & 3 & 90.6\% & 6 & 0.00051 & 88.375\% & 9\\
    \hline
    \multirow{2}{*}{2} & 4 & 88.625\% & 4 & 0.00070 &  87.875\%    &  9   \\
    & & & & & & \\[-1em]
    & 3 & 88.75\% & 9 & 0.00037 & 89.375\% & 8\\
    \hline
    \multirow{2}{*}{3} & 4 & 86.875\% & 7 & 0.00043  &  87.375\%    &  8   \\
    & & & & & & \\[-1em]
    & 3 & 89.5\% & 9 & 0.00118 & 87.75\% & 7\\
    \hline
    \multirow{2}{*}{4} & 4 & 80.0\% & 3 & 0.00107 &  83.0\%   &  8   \\
    & & & & & & \\[-1em]
    & 3 & 80.5\% & 8 & 0.00045 & 90.75\% & 9\\
    \hline
    \multirow{2}{*}{5} & 4 & 88.375\% & 9 & 0.00080  &  87.0\%   &  9  \\
    & & & & & & \\[-1em]
    & 3 & 90.75\% & 9 & 0.00040 & 87.75\% & 9\\
    \hline
    \multirow{2}{*}{6} & 4 & 79.375\% & 8 & 0.00079  &  83.875\%    &  6   \\
    & & & & & & \\[-1em]
    & 3 & 84.125\% & 8 & 0.00051 & 87.75\% & 9 \\
    \hline
    \multirow{2}{*}{7} & 4 & 79.875\% & 3 & 0.00259  &  85.375\%    &  9   \\
    & & & & & & \\[-1em]
    & 3 & 85.625\% & 8 & 0.00071 & 89.25\% & 9\\
    \hline
    \multirow{2}{*}{8} & 4 & 81.75\% & 6 & 0.00098  &  90.0\%    &  9   \\
    & & & & & & \\[-1em]
    & 3 & 94.875\% & 8 & 0.00083 & 90.125\% & 8\\
    \hline
    \multirow{2}{*}{9} & 4 & 82.25\% & 9 & 0.00102  &  85.0\%    &  8   \\
    & & & & & & \\[-1em]
    & 3 & 85.0\% & 8 & 0.00076 & 84.75\% & 4\\
    \hline
    \multirow{2}{*}{10} & 4 & 80.375\% & 5 & 0.00079  & 87.0\%    &  9   \\
    & & & & & & \\[-1em]
    & 3 & 92.0\% & 9 & 0.00077 & 87.25\% & 9\\
    \hline
    \multirow{2}{*}{11} & 4 & 72.278\% & 1 & 0.00070  &  88.734\%    &  9   \\
    & & & & & & \\[-1em]
    & 3 & 82.152\% & 9 & 0.00032 & 87.468\% & 9\\
    \hline
     \multirow{2}{*}{Average} & 4 & 83.388\% & 6 & 0.00098 & 86.749\% & 8.17\\
     & & & & & & \\[-1em]
     & 3 & 87.613\% & 8.25 & 0.00061 & 88.237\% & 8.25\\
    \hline

    \end{tabular}

    \end{center}
\end{table*}

Models with 3 layers outperformed on average on both accuracies of interest as well as MBM Validation Loss, so we proceeded to finetuning with the saved 3-layer models. The exact details of finetuning dataset construction can be found in the Materials section below, but as in the pretraining phase, each fold has 10,000 training samples and 800 validation samples.

\subsection{Finetuning}
We loaded the twelve 3-layer models saved after their Best Epoch during the Multitask and NTP-only regimens and trained them for ten epochs on the Same Genre task described above. The training data for each model holds out the same run as was held out during pretraining as a validation split. Preliminary testing showed that freezing the pretrained weights and updating only the new output block was not a successful training strategy for this work. Therefore all parameters were updated during finetuning. To examine the benefit of transfer learning, we also trained twelve ``fresh" models. The fresh models are identical to the other models used in finetuning but do not load any pretrained weights.

The Adam optimizer parameters were the same as during pretraining. We trained all 36 models with a Learning Rate of $10^{-4}$ and then again with $10^{-5}$- the two learning rates used during pretraining. Table~\ref{table:finetunecrossval} gives the $10^{-5}$ results. These results outperformed the $10^{-4}$ results across the board so those are not reported.

\begin{table*}[t]
    \begin{center}
    \caption{Results of 12-fold cross validation for three finetuning regimens on the Same Genre task: Multitask-pretrained models, NTP-only-pretrained models, and fresh models. Best Val Acc is the highest accuracy obtained during training on validation split. The epoch in which that accuracy was obtained is given in the Best Epoch column, from 0 to 9 inclusive. The average across all twelve folds is given at the bottom of each column.}
    \label{table:finetunecrossval}
    \begin{tabular}{|| m{2cm}| m{2cm} | m{2cm} | m{2cm} | m{2cm} | m{2cm} | m{2cm} ||}
    \hline
    & \multicolumn{2}{|c|}{\textbf{Multitask (NTP and MBM)}}& \multicolumn{2}{|c|}{\textbf{NTP Only}} & \multicolumn{2}{|c||}{\textbf{Fresh}}\\
    \hline
    \textbf{Heldout Run} & \textbf{Best Val Acc} & \textbf{Best Epoch} & \textbf{Best Val Acc}   & \textbf{Best Epoch}  & \textbf{Best Val. Acc} & \textbf{Best Epoch}\\
    \hline
    0 & 82.625\% & 7 & 94.75\% & 9  &  84.625\%   &  8   \\

    \hline
    1 & 86.625\% & 9 & 91.375\% & 9  &  88.75\%   &  5   \\

    \hline
    2 & 88.375\% & 9 & 94.0\% & 6 &  89.625\%    &  6   \\

    \hline
    3 & 93.0\% & 4 & 92.125\% & 8  &  89.5\%    &  9   \\

    \hline
    4 & 72.75\% & 9 & 93.25\% & 8 &  82.5\%   &  6   \\

    \hline
    5 & 89.5\% & 9 & 92.0\% & 5  &  86.5\%   &  9  \\

    \hline
    6 & 82.0\% & 9 & 91.0\% & 9  &  82.75\%    &  9   \\

    \hline
    7 & 98.25\% & 9 & 90.875\% & 5  &  82.5\%    &  9   \\

    \hline
    8 & 94.25\% & 9 & 95.375\% & 9  &  83.625\%    &  6   \\

    \hline
    9 & 78.125\% & 9 & 92.375\% & 9  &  83.125\%    &  5   \\

    \hline
    10 & 82.625\% & 7 & 91.875\% & 8  & 87.875\%    &  8   \\

    \hline
    11 & 81.392\% & 2 & 97.089\% & 8  &  83.291\%    &  9   \\

    \hline
     Average & 85.793\% & 7.67 & 93.007\% & 7.75 & 85.389\% & 7.42\\

    \hline

    \end{tabular}

    \end{center}
\end{table*}

\subsection{Discussion}
The first point of interest is that the pretraining phase was successful at all. fMRI data is a challenging domain and paired-sequence transformers have not previously been used in this domain, nor has multitask learning, in addition to our pretraining tasks being novel. Nevertheless, our implementation is conclusively capable of learning these tasks. The average best performance between the two regimens is not significantly different (87.6\% vs. 88.2\%), which alleviates concerns about MBM impeding the ability to learn NTP. Moreover, it does not impede the \textit{speed} at which the multitask models achieve their best performance- about 8 epochs in both cases. The multitask models are more volatile, with lower lows but also higher highs. NTP-only achieves a highest validation accuracy of 90.75\%, but multitask runs achieve 92\%, 93.5\%, and 94.875\%, which is \textbf{our first evidence of a synergistic benefit from self-supervised multitask training on fMRI data.}  

Our novel supervised brain decoding task, Same Genre, was also successful on both pretrained models and fresh models. The models pretrained on NTP-only significantly outperformed the fresh models, which is \textbf{our first significant evidence of the ability to perform transfer learning with our model from one of our novel self-supervised pretraining tasks to a supervised brain decoding task.} The models pretrained on Multitask almost exactly matched the baseline fresh models on average, but we note a similar volatility to the pretraining phase. The average of the Multitask models is dragged down by folds 4 (72.75\%) and 9 (78.125\%). On the other hand, fold 7 achieves a staggering 98.25\% validation accuracy, as well as 93\% and 94.25\%, all of which exceed the fresh models' best fold of 91.625\%. NTP-only reached a maximum of 97.089\%, which is also short of the Multitask maximum.

The relationship between pretraining performance and finetuning performance is unclear. For example, the second highest finetuning accuracy for Multitask was on folds 8, which was the highest performance of the 3-layer models during pretraining, indicating the positive relationship between the two phases that we would expect. On the other hand, fold 7 had the best performing Multitask finetuning accuracy, or rather the best finetuning accuracy of any regimen, while the pretraining accuracy and MBM loss were both below average on that fold. More work is required to properly identify a relationship between the two phases.

\section{Conclusions and Future Work}
In this work we presented two novel self-supervised tasks for pretraining on sequences of fMRI data- Next Thought Prediction, and Masked Brain Modeling. The results of our pretraining phase demonstrated our paired-sequence transformer architecture's successful learning of those tasks, as well as the first evidence of a synergistic benefit from multitask training on fMRI data. The results of our finetuning phase demonstrated transfer learning from NTP to a supervised brain decoding task, establishing a proof of concept of our framework's suitability for transfer learning on fMRI data, in particular, in the absence of an embedding layer.

Observe the similarity between the Same Genre task and a standard contrast analysis of genre in STG. Fundamentally, both are asking if STG encodes different genres in different ways. Our implementation answers this question in the affirmative without the need for a contrast analysis. More generally, any contrast analysis interested in the difference between two conditions could be substituted by our model. Our upcoming work aims to show that our model can detect such differences which univariate approaches, such as contrast, fail to recognize.

Looking elsewhere, the experiments in this work do not exploit a particular power of transformers for learning long term dependencies. Increasing the sequence length would of course reduce the size of the training set, and this was our primary motivation for using a small sequence length in these experiments, but there were also some restrictions in the data collection protocol preventing us from wielding longer sequences. This is discussed in detail in the Materials section below. The Human Connectome Project (2013) \cite{hcp} is a prevailing candidate for future on long sequences of fMRI data due to its overwhelming size and well-established benchmarks. 

As stated above, we believe this paired-sequence framework has potential for replacing or supplementing contrast analysis, but there is also a great deal of work with fMRI data, for example basic brain decoding, that cannot use the paired-sequence structure. Our upcoming work has a novel self-supervised pretraining task for only \textit{one} sequence, which generalizes immediately to common brain decoding tasks and datasets.

\section{Materials}
\subsection{Data Preprocessing} 
The training data for this work were built from the Music Genre fMRI Dataset (2021) \cite{opengenre} available on the OpenNeuro database. This dataset contains whole-brain images of five subjects listening to 540 music pieces from 10 music genres. We refer to the original paper \cite{nakai2021} for the full details of data collection, but much of it has been covered in this work as it became relevant. We used the Brainlife (2017) \cite{brainlife} application to conduct our suite of preprocessing for the Music Genre dataset. The fMRIprep (2019) \cite{fmriprep} preprocessing package performed motion correction, field unwarping, normalization, bias field correction, and brain extraction. Next, fMRIprep mapped the 3-D images of each participant's brain into a standardized 3-D (96x114x96) vector space (MNI space) \cite{MNIspace} such that the various structural components of each brain are at the same coordinates. This enables direct comparison and analysis of BOLD signals in physically different brains. In our case, MNI space facilitates among-subject training of machine learning models. We refer to this standardized vector space throughout the paper as the \textbf{voxel space}, where the 3-D coordinates correspond to $1x1x1mm^3$ cubes of brain called \textbf{voxels} and the value at each coordinate is the BOLD signal in that volume. To summarize, the \textbf{voxel data} used in model training is the timeseries of 3-D BOLD signals after preprocessing and mapping to MNI space. (Fig. \ref{fig:voxels}). 

\subsection{Region of Interest}
The full MNI space is several orders of magnitude too large for our purposes, but more importantly we are only interested in regions of the brain recruited during attentive listening to music. The Superior Temporal Gyrus (STG) is the site of the auditory cortex, which processes auditory information. Angulo-Perkins et al. (2014) \cite{nonmusicians} showed preferential involvement of STG in processing music in both musicians and non-musicians, which fits our goal of learning from the Music Genre dataset. STG has also been used to learn decoding models of complex natural sounds \cite{complexfmri}, language \cite{languagefmri}, and even imagined sound \cite{imagined} from fMRI data. Therefore we chose to extract STG from each subject's voxel data. 
\begin{figure*}[t]

    \centering
    \includegraphics[width=0.5\textwidth]{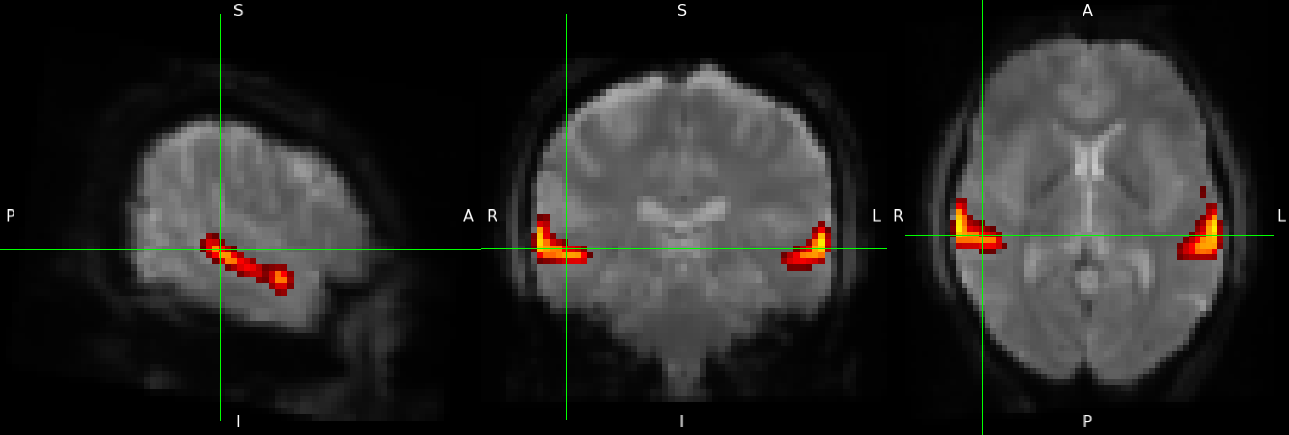}
    \caption{Heatmaps for probability of voxel inclusion in STG. Only probabilities greater than or equal to 23\% are shown.}
    \label{fig:STGheatmap}
\end{figure*}
\begin{figure*}[t]

    \centering
    \includegraphics[width=0.5\textwidth]{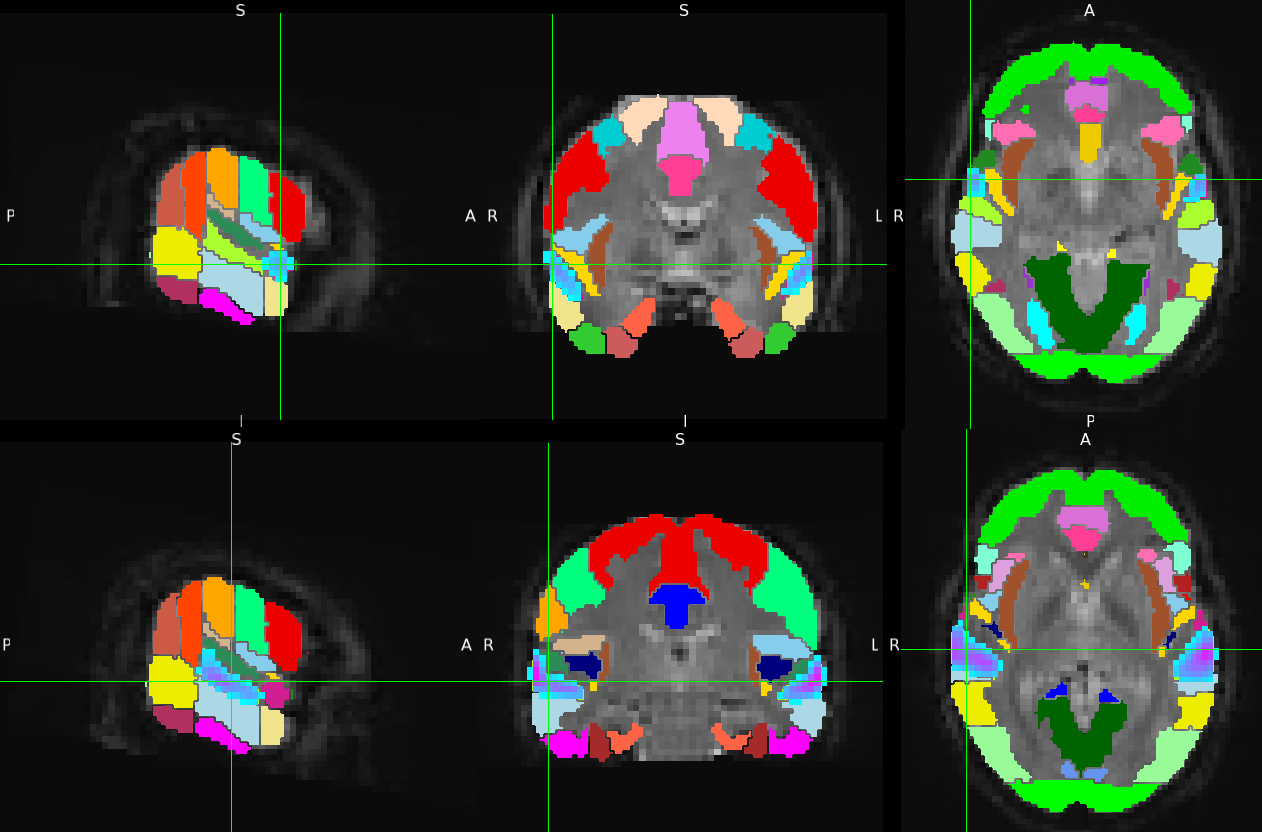}
    \caption{Heatmaps for probability of a particular voxel being included in a chosen region of interest. Anterior STG (top) and Posterior STG (bottom) maps with a minimum inclusion probability of 23\% are overlayed on the Harvard-Oxford atlas.}
    \label{fig:harvardoxford}
\end{figure*}
FSLEyes (2022) \cite{fsleyes} is a free application for viewing fMRI images and includes several \textbf{atlases} for isolating structural ROIs in the brain with respect to MNI space. We used the Harvard-Oxford cortical structure atlas (HO atlas), some regions of which are shown as an example in Figure \ref{fig:harvardoxford}. 

The HO atlas assigns a probability to each voxel of belonging to each ROI. Therefore in order to extract STG, we needed to choose a minimum probability threshold for inclusion in STG. This threshold is a hyperparameter to be tuned in future work, but in this work all datasets have a threshold of 23\%. We obtained our threshold by visual inspection of the resulting regions. In their seminal work, Craddock et al. (2011) \cite{craddock2011} used a threshold of 25\% with the HO atlas, so our visual inspection method is only slightly more lenient. 

The HO atlas labels anterior and posterior STG separately, so we applied the threshold to both regions and concatenated them. Voxels which met the threshold for inclusion in \textit{both} anterior and posterior are only included once and the greater of the two probabilities is preserved. Figure \ref{fig:STGheatmap} shows the heatmap corresponding to this union.  

We proceed with only one of the two lateral STGs for reduced model complexity and thus lower resource demand for training. In our previous work \cite{audimg} we successfully decoded a particular auditory stimulus information from left-STG but not right-STG, and this was our primary basis for choosing left rather than right for these experiments. The number of voxels in left-STG with inclusion threshold $23\%$ is 413. However, to perform multi-head attention, the dimension of the input to a transformer must be evenly divisible by the number of attention heads. Thus our inputs should have a dimension (i.e. number of voxels) with a ``nice'' factorization to evaluate different numbers of attention heads. The choice was between \textit{removing} the voxels with the \textit{lowest} probabilities of inclusion \textit{above} $23\%$ to reach 400, or inserting voxels with the highest probabilities below $23\%$ to reach 420. We chose the latter for its more diverse factorization, that is, allowing for all of 2, 3, 4, 5, 6, and 7 attention heads. We then performed linear detrending across the full scan to each voxel and, finally, standardized each voxel to mean zero across the full scan. 

\subsection{Creating the Training Data}
The original dataset was collected by scanning while 15-second clips of music played with a 1.5 second repetition time (TR). So each clip corresponds to 10 consecutive images in the dataset. There were no rest periods between clips. Therefore, with the Same Genre task in mind, the most natural sequence length for each of of the inputs to the model is 10, but we chose 5 for more training data. Thus each sequence of 5 (5-seq) for the SG task is either timesteps 1 through 5 or 6 through 10 of a music clip. All 5-seqs were extracted. For each 5-seq, a positive and negative training sample was created. That is, each 5-seq was inserted to the left of the SEP token, and a 5-seq with the same music genre was sampled uniformly at random \textit{from the same subject's data} and placed to the right of the SEP token to create a positive sample. Similarly for a negative sample. We have not yet trained this task when the pairs are among-subject, but the models in this work are trained on the collection of pairs from all subjects, resulting in a within-among-subject hybrid. The size of our SG datasets could be multiplied by making more positive and negative samples, but that would of course multiply the training time, and our results were already strong on this task.

Recall that the masking process for MBM is done when an input arrives at the model, so creating a dataset for multitask training reduces to creating a dataset for NTP-only. We needed to use 5-seqs to make training data for the NTP task to have the correct model specifications for finetuning on SG. All 5-seqs were extracted and both a positive and negative within-subject sample were created, as with SG. These 5-seqs could have arbitrary start and end points and possibly overlap, but we chose the same start and end points as for SG. This was because it sets up an interesting challenge for the model to overcome. When the 5-seq is images 1 through 5 for a music clip, the positive sample for NTP is necessarily a positive sample for SG. But when the 5-seq is images 6 through 10, the positive sample for NTP is necessarily a negative sample for SG- that is, the 5-seq immediately following it must be for a different genre (a genre never followed itself during scanning). In simple terms, for half the samples where ``yes" is the correct answer during pretraining, ``no" would be the correct answer during finetuning. Thus this construction presented a meaningful hurdle for the models to overcome during finetuning, they could not simply follow the same pathways they learned during pretraining.

The training data for each fold of the cross-validation was constructed by holding out one of the twelve runs labelled ``Training Run" in the original dataset. The inputs to the model contain two 5-seqs, so inputs in the validation splits were constructed by sampling both 5-seqs from the heldout run, and the training splits did not sample from the heldout runs. Note that while the dataset construction process was the same for pretraining and finetuning, the datasets themselves are \textit{not} the same. We constructed a new dataset for each fold of pretraining and finetuning. Within each fold of pretraining, Multitask and NTP-only used the same dataset in order to directly compare performance. Similarly, for each fold of finetuning, the three regimens used the same dataset.

In addition to the twelve runs labelled ``Training Run" in the original dataset, there are six ``Test Runs," which were constructed slightly differently. We emphasize here that the words ``training" and ``test" in the original run labels have no relation to our own training and validation splits. Each ``Training Run" corresponds to 40 different music clips, while each ``Test Run" corresponds to a sequence of 10 music clips (one from each genre) repeated four times. When creating our datasets, we only included 5-seqs from the \textit{first} instance of each clip in the ``Test Runs," and the others were discarded. The result for each fold was 10000 total training samples and 400 validation samples.

Finally, recall that our architecture does not have an embedding layer. In NLP, the tokens are added as word indices to the vocabulary and the embedding layer learns their distributed representations \cite{vaswani2017}. Malkiel et al. (2022) \cite{malkiel} prepend a CLS token to sequences of fMRI images and pass that sequence through a learned embedding layer. But the original form of the CLS token must have the same dimension as the fMRI images in the sequence in order for the embedding layer to accept it. They do not report what this original form was. Logically, the tokens ought to be ``far away" from the rest of the data in the distributed space. Thus we simply reserved the first three of the 420 dimensions for our tokens. The CLS, SEP, and MSK tokens have a $1$ in the zeroth, first, and second dimensions respectively, and are zero elsewhere. Each fMRI image has zero in those dimensions. Indeed, we had to remove the three voxels with lowest probability of inclusion from each image to make room for the tokens, and thus in practice only had 417 voxels in each image rather than 420. The success of our training validates our implementation of these tokens without an embedding space.

\ifCLASSOPTIONcaptionsoff
  \newpage
\fi



%
\nocite{*}

\bibliographystyle{IEEEtran}
\bibliography{pairedinputs}



\end{document}